\documentclass{article}

\PassOptionsToPackage{numbers}{natbib}
\usepackage[preprint]{neurips_2022}

\usepackage[utf8]{inputenc} % allow utf-8 input
\usepackage[T1]{fontenc}    % use 8-bit T1 fonts
\usepackage{hyperref}       % hyperlinks
\usepackage{url}            % simple URL typesetting
\usepackage{booktabs}       % professional-quality tables
\usepackage{amsfonts}       % blackboard math symbols
\usepackage{nicefrac}       % compact symbols for 1/2, etc.
\usepackage{microtype}      % microtypography
\usepackage{xcolor}         % colors
\usepackage[pdftex]{graphicx}
\usepackage{amsmath}
\usepackage{makecell}
\usepackage{multirow}
\usepackage{caption}
\usepackage{subcaption}
\usepackage{wrapfig}

\title{\texttt{VQ-AR}: Vector Quantized Autoregressive  Probabilistic Time Series Forecasting}

\author{%
Kashif Rasul\\
\texttt{kr95698@navercorp.com} \\
\And
Young-Jin Park\\
NAVER CLOVA \& NAVER AI Lab\\
\texttt{young.j.park@navercorp.com}\\
\And
Max Nihlén Ramström\\
NAVER CLOVA\\
\texttt{max.nihlen.ramstrom@navercorp.com}
\And
Kyung-Min Kim\\
NAVER CLOVA \& NAVER AI Lab\\
\texttt{kyungmin.kim.ml@navercorp.com}
}

\begin{document}

\maketitle

%%
%% The abstract is a short summary of the work to be presented in the
%% article.
\begin{abstract}
  Time series models aim for accurate predictions of the future given the past, where the forecasts are used for important downstream tasks like business decision making. In practice, deep learning based time series models come in many forms, but at a high level learn some continuous representation of the past and use it to output point or probabilistic forecasts. In this paper, we introduce a novel autoregressive architecture, \texttt{VQ-AR}, which instead learns a \emph{discrete} set of representations that are used to predict  the future. Extensive empirical comparison with other competitive deep learning models shows that surprisingly such a discrete set of representations gives state-of-the-art or equivalent results on a wide variety of time series datasets. We also highlight  the shortcomings of this approach, explore its zero-shot generalization capabilities, and  present an ablation study on the number of representations. The full source code of the method will be available at the time of publication with the hope that researchers can further investigate this important but overlooked  inductive bias for the time series domain.
\end{abstract}

\section{Introduction}
Time series forecasting is an important business and scientific machine learning problem which is typically used to support decision making for down-stream tasks. Classical methods like in~\cite{hyndman2021forecasting}   model each time series in a dataset individually using hand-crafted features and are used extensively as they providing strong baselines.

Recently deep learning based methods offer an alternative approach  by utilizing a shared model trained over the whole dataset of time series,  without the need for explicit feature engineering. These methods are competitive, especially when dealing with a large amount of data, against the classical methods, but  can also suffer from overfitting or learn spurious correlations due to their flexibility.

Often the forecasts are point valued, however it is much more useful for the decision making process to also incorporate the inherent  (aleatoric)  uncertainty present in the observations via \emph{probabilistic forecasts}. In the univariate setting this is usually done by learning the parameters of some chosen distribution, or via quantile regression or non-parametric and semi-parametric models which explicitly learn the conditional quantiles of the next time step given the past \cite{park2021learning}.  Other approaches include using Conformal Predictions \cite{2021conformal, pmlr-v139-xu21h}. In the multivariate setting \cite{tsay} however, one needs to resort to some approximation of the full multivariate distribution \cite{NIPS2019_8907} due to computational tractability or learn the conditional distribution via Normalizing Flows \cite{temp-flow, NEURIPS2020_1f47cef5}, Energy Based Methods \cite{pmlr-v139-rasul21a}, or Generative Adversarial Networks (GANs) \cite{engproc2021005040}.

As in most supervised deep learning methods, these models learn some representation of the history of a time series in order to best forecast, which can be thought of as a kind of self-supervised learning  task (not unlike \texttt{GPT} \cite{NEURIPS2020_1457c0d6} or \texttt{Bert} \cite{devlin-etal-2019-bert} in the Natural Language Processing (NLP) setting). Thus in order to provide better historical representations we have a large collection of architectural choices we can make for the problem or dataset  at hand. 

The core insight of this work is the realization that there might  only be a discrete collection of representations of time series histories which could potentially be used for the prediction task. The success of modern NLP on the back of discrete token representations leads us to consider if something similar would also work in the time series domain. Thus, in this paper  we take inspiration from the Vector Quantized-Variational AutoEncoder (VQ-VAE)~\cite{NIPS2017_7a98af17} model to learn  a set of \emph{latent} representations in order to best forecast in an autoregressive fashion.

Thus, the  main contribution of this paper is a novel autoregressive time series forecasting method which is trained end-to-end and provides excellent inductive bias for the probabilistic forecasting task. We highlight the model's performance and properties via extensive experimentation and show that it achieves comparable results against ensembles of tree-based methods without extensive feature engineering.

% The paper is organized as follows: we provide some introductory background in section \ref{sec:background}, followed by an exposition of the model in section \ref{sec:model} as well as the experiments in section \ref{sec:exp}. We then provide some related material in section \ref{sec:rel} and end with a summary and potential future work in section \ref{sec:summ}.

\section{Background} \label{sec:background}

We introduce some notation and go over some background material needed for the model first. 

In this work we assume the univariate forecasting setup of \texttt{DeepAR}~\cite{DBLP:journals/corr/FlunkertSG17}, however the method also applies to the multivariate setting. Formally we assume we are given a training dataset of $D \geq 1$ time series $\mathcal{D}_{\mathrm{train}} = \{x_{1:T_i}^i\}$ where $i \in \{1, \ldots, D\}$ and at each time $t$ we have $x_t^i \in \mathbb{R}$ or in $\mathbb{N}$.  The problem will require us to predict $P \geq 1$ steps into the future and thus will come with a back-testing test set of these $D$ time series denoted by  $\mathcal{D}_{\mathrm{test}} = \{x_{T_{i}+1:T_i+P}^i\}$. Note that even though we are denoting the time index $t$ by an incremental counter here, in reality each $t$ has a date-time associated with it, which increments regularly based on the frequency of the dataset. Often we will have all the $T_i$ be the same date-time for the time series of a particular dataset we use.

\subsection{Probabilistic Forecasting} \label{sec:prob-forecast}

In the  probabilistic forecasting problem, we wish to learn the unknown future distribution of the data given its past. Rather than considering the whole history of each time series $i$ in a dataset (which  might not be of the same size) we can instead consider some fixed context window of size $C \geq 1$ of our choosing and try to learn this potentially complex unknown distribution of the future values given the context window sized past denoted by
\begin{equation}\label{eqn:dist}
    p_{\mathcal{X}}(x_{T_i+1:T_i+P}^i | x_{T_i+1-C:T_i}^i).
\end{equation}
Thus, if we denote the parameters of our deep learning model by $\theta$, we can approximate (\ref{eqn:dist}) by  an autoregressive model which we can write via the chain-rule of probability as
$$p_{\mathcal{X}}(x_{T_i+1:T_i+P}^i | x_{T_i+1-C:T_i}^i; \theta) = \Pi_{t=T_i+1}^{T_i +P} p_{\mathcal{X}}(x_{t}^i | x_{T_i+1-C:t-1}^i; \theta).$$

And as we can see at a high level the probabilistic forecasting problem reduces down to learning some representation of the context window past together with a distribution model of the next time step(s) given this representation. For example, the \texttt{DeepAR} model uses a Recurrent Neural Network (RNN) (like the LSTM \cite{6795963} or GRU \cite{cho-etal-2014-properties}) to represent the context window history together with associated covariates, denoted by  real-valued $\mathbb{R}^F$ sized vectors $\mathbf{c}^i_{1:T_i+P}$ (known for all times), as
$$
\mathbf{h}_t = \mathrm{RNN}(\mathtt{concat}(x^i_{t-1}, \mathbf{c}^i_t), \mathbf{h}_{t-1}; \theta),
$$
with $\mathbf{h}_t \in \mathbb{R}^H$ and $\mathbf{h}_0 = \vec{0}$, to learn $p_{\mathcal{X}}(x^i_t | \mathbf{h}_t; \theta).$ The distribution head in \texttt{DeepAR} learns the parameters of some chosen distribution, e.g. the mean and variance of a Gaussian and we can maximize the log-likelihood of the distribution with respect to the ground-truth $x^i_t$ via SGD for all $i$ and $t$ in $\mathcal{D}_{\mathrm{train}}$.

By incorporating different architectural inductive biases to learn the historic representation together with different distribution heads, researchers have come up with a zoo of models to solve the probabilistic forecasting problem in the deep learning domain, for  not only the univariate and multivariate or regular and irregular time series setting, but also for spatial-temporal problems as well.

\subsection{VQ-VAE}

Leaving time series forecasting aside for the moment we can consider the fact that latent representations can potentially also be discrete. Thus instead of learning a continuous valued distributional representation as in  Variational AutoEncoders (VAE)~\cite{DBLP:journals/ftml/KingmaW19}, the VQ-VAE model learns a  fixed size or discrete set of representations of the data via an encoder-decoder bottleneck.

Formally the model consists of an encoder that maps  its input onto a fixed sized set of latent variables and a decoder that reconstructs the input from one of these fixed number of latents. Both the encoder and decoder use this shared codebook of vectors. Thus if we denote the encoding representation by $\mathbf{h}^{\mathrm{enc}} \in \mathbb{R}^E$, then this vector is quantized based on its distance to the prototype vectors in the codebook $\{ \mathbf{z}_1, \ldots, \mathbf{z}_J \}$ such that the $\mathbf{h}^{\mathrm{enc}} $ is replaced by the closest prototype vector:
\begin{equation}\label{eqn:quantize}
\mathrm{Quantize}(\mathbf{h}^{\mathrm{enc}}) := \mathbf{z}_n \quad \mathrm{where} \quad n = \mathop{\mathrm{arg\,min}}_{j} \| \mathbf{h}^{\mathrm{enc}} - \mathbf{z}_j \|_2,
\end{equation}
for a codebook size $J \geq 1$ of our choosing.

The mappings are learned by back-propagating the gradient of a reconstruction error via the decoder to the encoder using the Straight-Through gradient estimator~\cite{hinton2012, bengio2013estimating}. Apart from this loss the VQ-VAE model also has two terms that encourages the alignment of the vector space of the codebook with the output of the encoder, namely the codebook loss which pushes the selected codebook entry $\mathbf{z}$ closer to the encoder representation and the commitment loss which encourages the output of the encoder to stay close to the chosen codebook vector to stop it switching too frequently from one code vector to another.  The ``stop-gradient'' or ``detach'' operator $\mathtt{sg}(\cdot)$ blocks the gradients from flowing into its argument and  $\beta$ is the hyperparameter which penalizes the code vector fluctuating, then the two extra entries of the loss are given by
\begin{equation}\label{eqn:vq-loss}
    \| \mathtt{sg}(\mathbf{h}^{\mathrm{enc}}) - \mathbf{z} \|^2_2 + \beta \| \mathtt{sg}(\mathbf{z}) - \mathbf{h}^{\mathrm{enc}} \|^2_2.
\end{equation}

Since the optimal code would be the average of the representations, \cite{NIPS2017_7a98af17} presents  an exponential moving average update scheme of the latents instead of the codebook loss (first term of (\ref{eqn:vq-loss})) during training. The \texttt{SoundStream}~\cite{50491} paper  proposes to initialize the codebook  by the k-means centroids of the representations from the first batch. Additionally, as proposed in \texttt{Jukebox}~\cite{dhariwal2020jukebox}, we deploy the heuristic where we replace codebook vectors that have an exponential moving average cluster size less than some threshold $Q$ by a random vector from the batch.  

\section{\texttt{VQ-AR} Method} \label{sec:model}

We motivate this method with the observation that time series data in a dataset  visually looks similar with repeated patterns, and thus we ask if  instead of a continuous representation of the histories as learned by  most deep learning models, could a discrete set of representations suffice? With this in mind, we introduce the \texttt{VQ-AR} model which incorporates an encoder-decoder RNN together with a Vector Quantizer (VQ) bottleneck. 

\begin{figure}
  \centering
  \includegraphics[width=0.6 \textwidth]{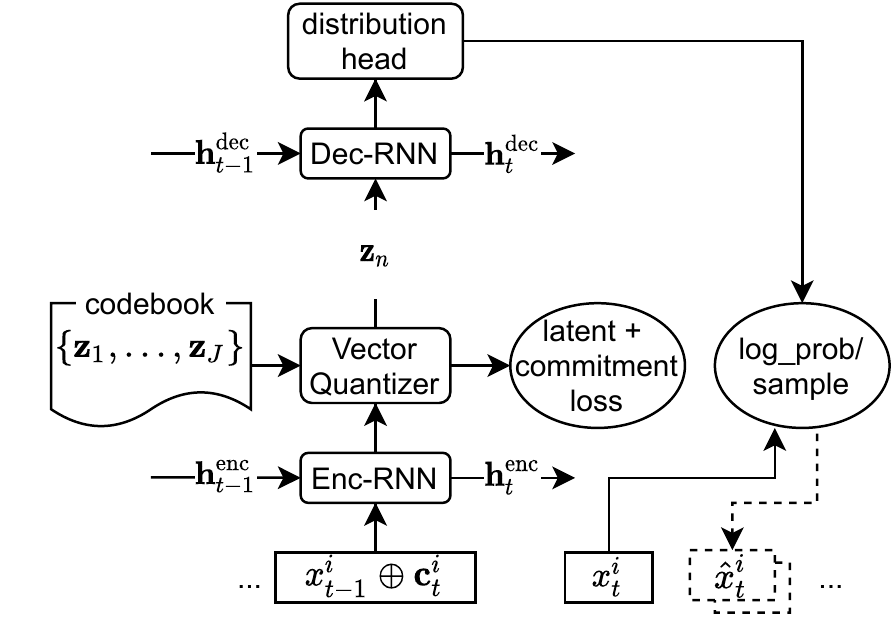}
  \caption{\texttt{VQ-AR} model schematic at time point $t-1$ for time series $i$. During training the model takes as input the target $x^i_{t-1}$ and covariates $\mathbf{c}^i_t$ and outputs the parameters of some chosen distribution to minimize the negative log-likelihood and Vector Quantizer loss (\ref{eqn:vq-ar-loss}); and during inference (dashed) it allows for sampling of the next time point $\hat{x}^i_t$, in an autoregressive fashion.}
  \label{fig:vq-ar}
\end{figure}

Unlike in the \texttt{DeepAR} setting,  where we learn the temporal representation of each time series' context window to predict the distribution of the next step, we  instead first encode it to a quantized representation $\mathbf{z}_n$ given by (\ref{eqn:quantize}) using an encoding-RNN's state denoted by
$$
\mathbf{h}_t^{\mathrm{enc}} = \mathrm{RNN}_{\mathrm{enc}}(\mathtt{concat}(x^i_{t-1}, \mathbf{c}^i_t), \mathbf{h}_{t-1}^{\mathrm{enc}}; \theta),
$$
with $\mathbf{h}_t^{\mathrm{enc}} \in \mathbb{R}^E$ and $\mathbf{h}^{\mathrm{enc}}_0 = \vec{0}$. Subsequently we model the distribution of the next time step $p_{\mathcal{X}}(x^i_t | \mathbf{z}_n; \theta)$ using a decoding-RNN
$$
\mathbf{h}_t^{\mathrm{dec}} = \mathrm{RNN}_{\mathrm{dec}}( \mathrm{Quantize}(\mathbf{h}^{\mathrm{enc}}_t), \mathbf{h}_{t-1}^{\mathrm{dec}}; \theta),
$$
which \emph{only} takes the Vector-Quantized latent $\mathbf{z}_n$  (\ref{eqn:quantize}) as input, with  $\mathbf{h}_t^{\mathrm{dec}} \in \mathbb{R}^H$ and $\mathbf{h}^{\mathrm{dec}}_0 = \vec{0}$.  See Figure~\ref{fig:vq-ar} for a schematic of the model.

Because the \emph{prior}  $p(\mathbf{z}_j) = 1/J$ is a uniform categorical distribution and the \emph{posterior} 
\begin{equation}
\label{eqn:posterior}
q(\mathbf{z}_n | x^i_{1:t-1}, \mathbf{c}^i_{2:t}; \theta) = 
\begin{cases}
1 & \text{if } n = \mathop{\mathrm{arg\,min}}_{j} \| \mathbf{h}_t^{\mathrm{enc}} - \mathbf{z}_j \|_2 \\
0 & \text{otherwise},
\end{cases}
\end{equation} 
is a categorical distribution parametrized by one-hot probability vectors, the KL-divergence term in the Evidence Lower Bound (ELBO)~\cite{bishop:2006:PRML} is constant ($\log J$), and since the ELBO's expectation term is over  a deterministic distribution (\ref{eqn:posterior}), we can simply train the whole model by maximizing the log-likelihood term of the next time step: $\log p_{\mathcal{X}}(x^i_{t} | \mathbf{z}_n, \mathbf{h}_{t-1}^{\mathrm{dec}}; \theta)$.

Thus, similar to  the \texttt{DeepAR} model we   minimize the negative log-likelihood of some chosen parametric distribution and train the whole model end-to-end together with the VQ loss (\ref{eqn:vq-loss}). For example, to model Gaussian emissions the distribution head consist of a linear layer which outputs the mean $\mu_t \in \mathbb{R}$ and standard deviation $\sigma_t \in \mathbb{R}^{+}$, making sure that the standard deviation is always positive via the $\mathtt{softplus}(\cdot)$ non-linearity. These parameters are used to instantiate  the distribution $\mathcal{N}(\mu_t, \sigma_t)$ from which we  calculate the negative log-likelihood: $- \log \ell_{\mathcal{N}}(x^i_t)$. The full loss of the model  for  time point $t-1$ in the training window and series $i$ is thus denoted by
\begin{equation}\label{eqn:vq-ar-loss}
    \mathcal{L}^i_{t-1}(\theta) := - \log \ell_{\mathcal{N}}(x^i_t)  + \|  \mathtt{sg}(\mathbf{h}^{\mathrm{enc}}_t) - \mathbf{z}_n \|^2_2 + \beta \| \mathtt{sg}(\mathbf{z}_n) - \mathbf{h}^{\mathrm{enc}}_t \|^2_2.
\end{equation}

\subsection{Training}

We  construct batches $\mathcal{B}$ of context and subsequent prediction  window of total size $T+1$,  by sampling a random time series (and corresponding covariates) from $\mathcal{D}_{\mathrm{train}}$ and then sample this window randomly within. We will then minimize the mini-batch loss, which is the mean of all the individual losses (\ref{eqn:vq-ar-loss}) denoted by
$$
\mathcal{L}(\theta) := \frac{1}{|\mathcal{B}| T} \sum_{x^i_{1:T+1} \in \mathcal{B}} \sum_{t=1}^T \mathcal{L}_t^i(\theta),
$$
via the Adam~\cite{kingma:adam} SGD optimizer at each step of training with an appropriate batch size $|\mathcal{B}|$ and learning rate.

\subsection{Inference}

At inference time we go over the last context sized window of each time series in the training dataset $\mathcal{D}_{\mathrm{train}}$ and then start forecasting for the $P$ prediction steps by sequentially sampling from the distribution head and autoregressively passing this value back into the model (together with the covariates which are known for all times). In fact, we can repeat this process $S$ number of times (e.g. $S=100$) to  then report any empirical probability interval of interest as well as probabilistic metrics, like Continuous Ranked Probability Score (CRPS)~\cite{doi:10.1198/016214506000001437, RePEc}, with respect to the ground-truth values in $\mathcal{D}_{\mathrm{test}}$.   Point forecasting metrics can also be evaluated with respect to the empirical median or mean from the $S$ samples at each time step of the prediction.

Note, unlike in some generative modeling use cases, we \emph{do not} sample with a reduced temperature distribution to obtain higher quality samples, nor do we  Beam-search or do Nucleus sampling as in NLP settings during inference.

\subsection{Covariates}

As mentioned in section \ref{sec:prob-forecast}, we can incorporate covariates for each time point of the time series (as well as for all future times) by creating time features given the granularity of the time series in question. For example, for daily data we could consider the day of the week, week of the month, month of the year, etc., features. Although not strictly necessary, but certainly helpful, we can  take inspiration from classical methods and construct time series covariates, such as lagged features, moving averages, difference to  previous time points, exponential moving averages, and moving percentiles, etc., of $x^{i}$ (and/or of temporal covariates) depending on the frequency of the time series in question.

Where applicable holidays or other recurring events can also be turned into temporal features. A running ``age'' counter can also serve as an indication of the length of a time series.
Apart from these temporal features, one can embed the categorical identity $i \in \{1, \ldots, D \}$ of each time series into a vector via a trainable embedding layer whose time independent output is copied over for all time points being considered. If the time series are grouped then their categorical tree structure can also be represented via similar embeddings.

Unless explicitly stated, for all our experiments we only use time, age,  lagged features, and categorical identities as covariates.

\subsection{Scaling} \label{sec:scaling}

Entities in time series data can potentially have arbitrary magnitudes and so to be able to train a shared deep learning model with respect to inputs of any scale we can utilize the following heuristic: for each context window we can calculate the mean value of the data in the interval, specifically $\nu^i = \sum_{t=1}^T x^i_t/ T$ (if it is not zero or $1$ otherwise) and divide the time series by this before using it in the model. The outputs from the distribution heads are then multiplied by this $\nu^i$  when sampling  or we rescale in the parameter space of the distribution (e.g. when we require integer samples from a Poisson or Negative-Binomial head). 

We  omit stating this heuristic in all our derivations and treat it as an implementation detail.

\section{Experiments} \label{sec:exp}

We test the performance of \texttt{VQ-AR} for the forecasting task in terms of performance and also highlight the limitations of the model and generalization properties of the representations in this section.

For our experiments we use \texttt{Exchange}~\citep{Lai:2018:MLS:3209978.3210006}, \texttt{Solar}~\citep{Lai:2018:MLS:3209978.3210006},  \texttt{Elec.}\footnote{\url{https://archive.ics.uci.edu/ml/datasets/ElectricityLoadDiagrams20112014}}, \texttt{Traffic}\footnote{\url{https://archive.ics.uci.edu/ml/datasets/PEMS-SF}}, \texttt{Taxi}\footnote{\url{https://www1.nyc.gov/site/tlc/about/tlc-trip-record-data.page}}, \texttt{Wikipedia}\footnote{\url{https://github.com/mbohlkeschneider/gluon-ts/tree/mv_release/datasets}} open datasets, preprocessed exactly as in~\cite{NIPS2019_8907}, as well as  \texttt{M5}~\cite{MAKRIDAKIS2021} and proprietary \texttt{E-Commerce} dataset, with their properties listed in Table~\ref{dataset} in the appendix \ref{sec:exp-detail}. As can be noted in the table, we do not need to normalize scales for the \texttt{Traffic} dataset. From the names of the datasets, we see that we cover a number of time series domains including finance, weather, energy, logistics, page-views, as well as E-commerce.

As indicated above, for evaluation we employ the CRPS metric which   measures the compatibility of a cumulative distribution function (CDF) $F$ with the ground-truth observation $x$ as
$$
    \mathrm{CRPS}(F, x) = \int_\mathbb{R} (F(y) -  \mathbb{I} \{x \leq y\})^2\, \mathrm{d}y,
$$
where $\mathbb{I}\{ x \leq y\}$ is the indicator function which is one if $x \leq y$ and zero otherwise. CRPS is a \emph{proper scoring function}, meaning CRPS attains its minimum when the predictive distribution $F$ and the data distribution are equal.
Employing the empirical CDF of $F$, i.e. $\hat F(y) = \frac{1}{S} \sum_{s=1}^S \mathbb{I}\{x^{(s)} \leq y\}$ using $S$ samples $x^{(s)} \sim F$ as a natural approximation of the predictive CDF, CRPS can be directly computed from  sampled predictions from the model at each time point~\citep{JSSv090i12}. The final metric is averaged over all the prediction time steps and time series in a dataset.

\subsection{Forecasting}

We will compare \texttt{VQ-AR} with the following  deep learning baseline \emph{probabilistic} univariate models
\begin{itemize}
    \item \texttt{DeepAR}~\cite{DBLP:journals/corr/FlunkertSG17}: an RNN based probabilistic model which learns the parameters of some chosen distribution for the next time point;
    \item  \texttt{MQCNN}~\cite{mqcnn}: a Convolutional Neural Network model which  outputs chosen quantiles of the forecast upon which we regress the ground truth via Quantile loss;
    \item \texttt{SQF-RNN}~\cite{pmlr-v89-gasthaus19a}: an RNN based non-parametric method which models the quantiles via linear splines and also regresses the Quantile loss;
    \item \texttt{IQN-RNN}~\cite{gouttes2021probabilistic}: combines an RNN model with an Implicit Quantile Network (IQN) \cite{pmlr-v80-dabney18a} head to learn the distribution similar to \texttt{SQF-RNN}; 
    \item \texttt{LSF}~\cite{hasson2021probabilistic}: a method that transform a point-estimator, coming from for example gradient boosting methods, into a probabilistic one; 
\end{itemize}
as well as the classical \texttt{ETS}~\cite{JSSv027i03}  which is an exponential smoothing method using weighted averages of past observations with exponentially decaying weights as the observations get older together with Gaussian \emph{additive} errors (E) modeling trend (T) and seasonality (S) effects separately. 

We evaluate the model on the  datasets detailed above and follow the recommendations of the M4 competition \cite{MAKRIDAKIS202054} regarding forecasting performance metrics. Thus, we also report the mean scale interval score \cite{doi:10.1198/016214506000001437} (MSIS\footnote{\url{http://www.unic.ac.cy/test/wp-content/uploads/sites/2/2018/09/M4-Competitors-Guide.pdf}}) for a 95\% prediction interval, the 50-th and 90-th quantile percentile loss (QL50 and QL90 respectively), as well as  the CRPS score. The point-forecasting performance of models is measured by the normalized root mean square error (NRMSE), the mean absolute scaled error (MASE) \cite{HYNDMAN2006679}, and the symmetric mean absolute percentage error (sMAPE) \cite{MAKRIDAKIS1993527}. For pointwise metrics, we use sampled \emph{medians} with the exception of NRMSE, where we take the \emph{mean} over our prediction samples. The results of our extensive experiments are detailed in Table~\ref{tab:metrics}. Note that the \texttt{VQ-AR} is essentially compressing the time series histories to discrete tokens which could have applications in edge-computing, by trading off performance with space.

\begin{table}
\caption{Comparison metrics using different methods: \texttt{SQF-RNN} with 50 knots, \texttt{DeepAR}  and  \texttt{VQ-AR} with Student-T (\texttt{-t}), Negative Binomial (\texttt{-nb}) or IQN (\texttt{-iqn}) emission heads, \texttt{ETS},  \texttt{MQCNN}, and \texttt{IQN-RNN} on the datasets.}
\label{tab:metrics}
\begin{center}

\begin{tabular}{lc|ccccccr}
\toprule
Dataset & Method & CRPS & QL50 & QL90 & MSIS & NRMSE & sMAPE & MASE \\
\midrule
\multirow{6}{*}{\texttt{Exchange}} 
 & \texttt{SQF-RNN-50} & 0.010 & 0.013 & 0.006 & \textbf{14.15} & 0.020 & 0.013 & 1.800\\ %done
 & \texttt{DeepAR-t} &  0.012 &  0.016 & 0.007 & 69.29  & 0.022  & 0.030  &  9.980  \\ %done
 & \texttt{ETS} & 0.008 & \textbf{0.010} & 0.005 & 15.89 & 0.015 & \textbf{0.011} & \textbf{1.517} \\
 & \texttt{IQN-RNN} & \textbf{0.007} &  \textbf{0.010} &  \textbf{0.004} &  17.37 & \textbf{0.014} & 0.013   & 3.041   \\
 & \texttt{MQCNN} & 0.015 & 0.016 & 0.011 & 60.04 & 0.026 & 0.045 & 5.440 \\
 & \texttt{\textbf{VQ-AR-t}} & 0.010 &  0.013  & 0.007  & 18.10  & 0.019 & 0.015  &  2.658 \\
 
 \midrule 
 
 \multirow{6}{*}{\texttt{Solar}} 
 & \texttt{SQF-RNN-50} & 0.330 & 0.431 & 0.175 & 5.65 & 0.929 & \textbf{1.342} & 1.004\\ %done
 & \texttt{DeepAR-t} &  0.418 & 0.543 &  0.254 &  7.33 &  1.072 &  1.393 & 1.275  \\ %done
 & \texttt{ETS} & 0.646 & 0.661 & 0.383 & 18.55 & 1.112 & 1.546 & 1.938 \\
 & \texttt{IQN-RNN} &  0.373 &  0.491 & 0.165 &  5.99 &  1.037 &  1.356 & 1.15 \\
 & \texttt{MQCNN} & 0.928 & 0.960 & 1.535 & 73.58 & 1.920 & 1.838 & 2.248 \\
 & \texttt{\textbf{VQ-AR-iqn}} & \textbf{0.320} & \textbf{0.414}  &  \textbf{0.174} &  \textbf{5.64} &  \textbf{0.885} & 1.346 &  \textbf{0.969} \\
 
 \midrule

\multirow{6}{*}{\texttt{Electricity}} 
  & \texttt{SQF-RNN-50}  & 0.078 & 0.097 & 0.044 &   8.66  &0.632 & 0.144 & 1.051 \\ % done
  & \texttt{DeepAR-t}  & 0.062 & 0.078 & 0.046 &  6.79 & 0.687 & 0.117 & 0.849 \\ %done
  & \texttt{ETS}  & 0.076 & 0.100 & 0.050 & 9.99 & 0.838 & 0.156 & 1.247  \\
  & \texttt{IQN-RNN}  & 0.060  & 0.074 & 0.040 &  8.74 &  \textbf{0.543} & 0.138 & 0.897 \\ %done
  & \texttt{MQCNN} & 0.129 & 0.148 & 0.132 & 30.54 & 1.230 & 0.240 & 2.000 \\
  & \texttt{\textbf{VQ-AR-t}} & \textbf{0.054} & \textbf{0.068} & \textbf{0.036} & \textbf{5.88} & 0.653 & \textbf{0.107} & \textbf{0.717}\\
  
\midrule

\multirow{6}{*}{\texttt{Traffic}} 
 & \texttt{SQF-RNN-50} & 0.153 & 0.186 & 0.117 & 8.40 & \textbf{0.401} &  0.243  & 0.76 \\ %done
 & \texttt{DeepAR-t} & 0.172 & 0.216 & 0.117 & 8.02 & 0.472  & 0.244 &0.89\\ %done
 & \texttt{ETS} & 0.373  & 0.386 & 0.287 & 17.67 & 0.647 &  0.489 & 1.543 \\
 & \texttt{IQN-RNN}  & 0.139 & 0.168 & 0.117 & \textbf{7.11} & 0.433 & \textbf{0.171} & 0.656 \\
 & \texttt{MQCNN} & 1.220 & 0.563 & 2.005 & 116.69 & 0.723 & 0.636  & 2.712\\
 & \texttt{\textbf{VQ-AR-t}} & \textbf{0.138} & \textbf{0.164} & \textbf{0.113} & 7.79 & 0.409 & 0.185 & \textbf{0.641} \\
\midrule

\multirow{6}{*}{\texttt{Taxi}} 
 & \texttt{SQF-RNN-50} & \textbf{0.286} & \textbf{0.362} & \textbf{0.188} & 5.53 & \textbf{0.570} & 0.609 &  \textbf{0.741}\\ %done
 & \texttt{DeepAR-nb} & 0.299  &0.379 &  0.203 & 5.44 &  0.610 & 0.582  &  0.771\\ %done
 & \texttt{ETS} & 1.059 & 1.297 & 0.617 & 12.24 & 2.147 & 1.159  & 1.552 \\
 & \texttt{IQN-RNN} & 0.295 & 0.370  &  0.201 & 6.51  & 0.583  &  0.629 & 0.758  \\
 & \texttt{MQCNN} & 1.262 & 1.451 & 0.488 & 48.61 & 2.645 & 0.912 & 3.041 \\
 & \texttt{\textbf{VQ-AR-nb}} & \textbf{0.286} & \textbf{0.362}  & 0.193 & \textbf{5.43}  &0.572 & \textbf{0.570} & \textbf{0.741} \\
 
 \midrule

\multirow{7}{*}{\texttt{Wikipedia}}
   & \texttt{SQF-RNN-50} &0.283 & 0.328 & 0.321 & 23.71 &  2.24  & 0.261 & 1.44 \\ %done
   & \texttt{DeepAR-nb}  &0.321  & 0.383 &  0.361  & 26.48  & 2.354 & 0.327  & 1.852 \\ %done
   & \texttt{DeepAR-t}  & 0.235 &0.27 &  0.267 & 23.77 & 2.15 & 0.219  & 1.295\\ %done
   & \texttt{ETS} & 0.788 & 0.440 & 0.836 & 61.68 & 3.261 & 0.301 & 2.214 \\
   & \texttt{IQN-RNN}  & \textbf{0.221} & \textbf{0.254} & \textbf{0.251}
   &  \textbf{21.78} & \textbf{2.102} & \textbf{ 0.193} &  \textbf{1.214} \\
   & \texttt{MQCNN} & 0.398 & 0.453 & 0.327 & 38.79 & 2.202 & 0.379 & 2.336 \\
   & \texttt{\textbf{VQ-AR-iqn}}  &  0.231 & 0.266 & 0.252 & 22.09 & 2.106 & 0.208 & 1.261 \\

\midrule
\multirow{6}{*}{\texttt{M5}} 
 & \texttt{SQF-RNN-50} & 0.558 & 0.708 & 0.466 & 8.26 & 1.634 & 1.556 & 0.912\\ %done
 & \texttt{DeepAR-nb} & 0.539  & 0.679  & 0.469 &  8.01 & 1.547  &  1.550 &  0.915   \\ %done
 & \texttt{ETS} & 0.838 & 1.051 & 0.696 & 23.67 & 2.560 & 1.560 & 3.161 \\
 & \texttt{IQN-RNN} &  0.539 &  \textbf{0.677} &  0.475 &  8.26 &  \textbf{1.511} & 1.602 &  0.898 \\
 & \texttt{MQCNN} & 0.574 & 0.725 & 0.497 & 8.04 & 1.775 & 1.616 & 0.921 \\
 & \texttt{\textbf{VQ-AR-nb}} & \bf  0.527 & 0.694 & \bf 0.457 & \bf 7.20  & 1.628 & \bf 1.535 & \bf 0.895   \\
%  & \texttt{\textbf{VQ-AR-nb}} & \textbf{0.531} & 0.697 & \textbf{0.466} & \textbf{7.24} &  1.649 & \textbf{1.539} & 0.903   \\
 
 \midrule

\multirow{6}{*}{\texttt{E-Commerce}} 
 & \texttt{SQF-RNN-50} & 0.545  & 0.627 & 0.648 & 24.82 & 8.492 & 1.653 & 1.246 \\ %done
 & \texttt{DeepAR-nb} & 0.531 & 0.612 & 0.638 & 38.12 &  8.160   & \textbf{1.409} & \textbf{1.071} \\ %done
 & \texttt{ETS} & 2.605 & 2.991 & 2.457 & 117.50 & 31.236 & 1.791 & 10.270 \\
 & \texttt{IQN-RNN} &  0.534  & 0.627  & 0.590 &  32.92 &  7.588 & 1.650 & 1.076 \\
 & \texttt{MQCNN} & 0.649 & 0.751 & 0.712 & 27.56 & 9.336 & 1.677 & 1.325 \\
 & \texttt{\textbf{VQ-AR-iqn}} & \textbf{0.483} & \textbf{0.562} & \textbf{0.555} & \textbf{22.33}  & \textbf{6.702} & 1.629 & 1.074 \\

\bottomrule
\end{tabular}
\end{center}
\end{table}

The M5 competition was a twin competition measuring both point forecasting accuracy \cite{MAKRIDAKIS2021} as well as probabilistic uncertainty \cite{MAKRIDAKIS2021-uncertainty}   and the top solutions were dominated by tree-based methods. In fact, the winning solution to the accuracy benchmark is based on an ensemble of 220 gradient boosting tree (GBT) models while the winner of the uncertainty challenge is an ensemble of 126 GBTs with hundreds of hand-crafted features. In Table~\ref{tab:m5} we provide a comparison between tree-based methods and neural methods. As detailed in~\cite{JANUSCHOWSKI2021} it is still an open problem to improve deep-learning models with respect to tree-based methods and as can be seen, the \texttt{VQ-VR} method provides competitive metrics via a simple \emph{single} model without any extensive feature engineering.

\begin{table}
  \caption{Weighted Quantile Loss (lower is better) for \texttt{M5} test set predictions using different methods for specified quantiles and their average.}
  \label{tab:m5}
  \begin{tabular}{l|ccccccc}
    \toprule
    Quantile & \makecell{\texttt{M5}\\ winner} & \makecell{\texttt{M5}\\ runner-up} & \texttt{DeepAR-t} & \texttt{DeepAR-nb} & \texttt{MQCNN} & \texttt{LSF} & \texttt{\textbf{VQ-AR-nb}}\\
    \midrule
0.005	& \textbf{0.010} & 0.042	&0.023	&0.016	&\textbf{0.010} & 0.012 &0.013\\
0.025	& \textbf{0.050} & 0.086  &0.075	&0.061	&0.051& 0.054 & 0.057\\
0.165	& \textbf{0.312} & 0.337	&0.319	&0.319	&0.320&  0.316 & 0.316 \\
0.25	& \textbf{0.444} & 0.461 &0.458	&0.450	&0.451&  0.451 & 0.445 \\
0.50	& 0.698 & 0.690	&0.766	&0.705	&0.712& 0.724 & \textbf{0.697} \\
0.75	&\textbf{0.687} & 0.722 &0.821	&0.699	&0.709& 0.728&  0.690 \\
0.835	& \textbf{0.585}	& 0.598 &0.754	&0.600	&0.609& 0.607 & 0.591 \\
0.975	&\textbf{0.182}	& 0.196 &0.430	&0.202	&0.213& 0.218 & 0.189 \\
0.995	&\textbf{0.055}	& 0.076 &0.310	&0.074	&0.084& 0.086 &0.063 \\
\midrule
  Mean & \textbf{0.336} & 0.357 & 0.440& 	0.347& 	0.351& 0.355 & 0.340 \\
  \bottomrule
\end{tabular}
\end{table}

\subsection{Robustness}

\begin{wraptable}{R}{0.5 \textwidth}
    \caption{Comparison of test set performance of \texttt{IQN-RNN} and \texttt{VQ-AR} on \texttt{Solar} dataset with different levels of Gaussian noise added to the context window input at the start of inference. The best metric for each noise level $l$ is highlighted in bold.}
\scalebox{0.85}{
\begin{tabular}{lc|cccc}
\toprule
Method & Level & CRPS & MSIS & sMAPE & MASE \\
\midrule
\multirow{6}{*}{\texttt{IQN-RNN}} 
&0.0 & 0.373 & 5.99 & 1.356 & 1.115 \\
&0.2 & 0.397 & \textbf{8.22} & \textbf{1.358} & 1.182\\ 
&0.4 & \textbf{0.431} & \textbf{11.97} & \textbf{1.364} & \textbf{1.234} \\
&0.6 & \textbf{0.479} & 17.49 & \textbf{1.370} & \textbf{1.324} \\
&0.8 & \textbf{0.536} & 22.71 & \textbf{1.382} & \textbf{1.453}\\
&1.0 & \textbf{0.610} & 29.42 & \textbf{1.399} & \textbf{1.612} \\
\midrule

\multirow{6}{*}{\texttt{\textbf{VQ-AR-iqn}}} 
&0.0 & \textbf{0.320} & \textbf{5.64} & \textbf{1.346} & \textbf{0.969} \\
&0.2 & \textbf{0.388} & 8.95 & 1.362 & \textbf{1.144} \\
&0.4 & 0.466 & 12.76 & 1.384 & 1.340 \\
&0.6 & 0.540 & \textbf{15.25} & 1.410 & 1.564 \\ 
&0.8 & 0.623 & \textbf{17.66} & 1.429 & 1.801 \\
& 1.0 & 0.712 & \textbf{21.18} & 1.451 & 2.042 \\
\bottomrule
\end{tabular}
}
\label{tab:robust}
\end{wraptable}

We check the robustness of \texttt{VQ-AR} with  respect to the noise added to the context window inputs (but not covariates) when forecasting. Our hypothesis is that due to the quantization nature of  the method, the added noise (up to a certain level) will correspond to the \emph{same} codebook vector and thus this method would be robust to this, as opposed to for example \texttt{DeepAR}. However, too much  noise could also cause the encoder to flip to nearby codebook vectors, and cause the predictions to be completely off  since there is no semantic meaning in nearby codebook vectors that is explicitly built into \texttt{VQ-AR}.

For some noise level $l = \{0.2, 0.4, \ldots, 1.0\}$ we calculate the standard deviation of the context window target $\sigma_{\textrm{context}}$  and \emph{add} to each time point noise sampled from $\mathcal{N}(0, l \times \sigma_{\textrm{context}})$ and then pass this together with the covariates to the appropriate RNN.

Table~\ref{tab:robust} details the metrics for different levels of noise. As we can see with a small amount of noise the \texttt{VQ-AR} method is more robust than \texttt{IQN-RNN}, however as we increase the noise the selected codebook vectors become too different leading to worse performance than with an RNN based method with smooth representations. The MSIS metric which measures the upper 95\% and lower 5\% probability interval however is more robust.

\subsection{Generalization}

The \texttt{VQ-AR} method is essentially an information bottleneck that serves as a form of regularization. Thus our hypothesis is that the learned cookbook representations $\{ \mathbf{z}_j \}$  would need to be somewhat universal time series representations in comparison to methods that can potentially learn continuous representations without such a constraint. In order to test this we measure the zero-shot prediction performance of \texttt{VQ-AR} against \texttt{DeepAR} by first training these two models on the \texttt{Electricity} dataset (without any categorical time series covariates) and then testing the trained model's performance on the test set of \texttt{Solar} and \texttt{Traffic}. 

\begin{wraptable}{R}{0.5\textwidth}
    \caption{Comparison of test set performance of \texttt{DeepAR} and \texttt{VQ-AR}  trained on the  \texttt{Electricity} dataset and then tested on \texttt{Solar}  and \texttt{Traffic} in a zero-shot setting. The best metric for each test dataset is given in bold.} 
\scalebox{0.77}{    
\begin{tabular}{lc|cccc}
\toprule
Method & Test set & CRPS & MSIS & sMAPE & MASE \\
\midrule
\multirow{2}{*}{\texttt{DeepAR-t}} 
& \texttt{Solar} & 0.522 & 16.60 & 1.441 & 1.432  \\
& \texttt{Traffic}  & 0.267 & 12.91 & 0.393 & 1.270\\ 

\midrule

\multirow{2}{*}{\texttt{\textbf{VQ-AR-t}}} 
&\texttt{Solar}  & \textbf{0.511} & \textbf{14.69} & \textbf{1.435} & \textbf{1.413} \\
&\texttt{Traffic}  & \textbf{0.214} & \textbf{10.52} & \textbf{0.300} & \textbf{0.972} \\
\bottomrule
\end{tabular}
}
\label{tab:gen}
\end{wraptable}

Table~\ref{tab:gen} highlights the prediction metrics on unseen datasets and as can be noted, the representations learned as discrete entities tend to perform better in the zero-shot setting especially since the datasets come from dissimilar domains.

\subsection{Ablation}

\begin{figure}
    \centering
    \begin{subfigure}[b]{0.45 \textwidth}
        \centering
        \includegraphics[width=\textwidth]{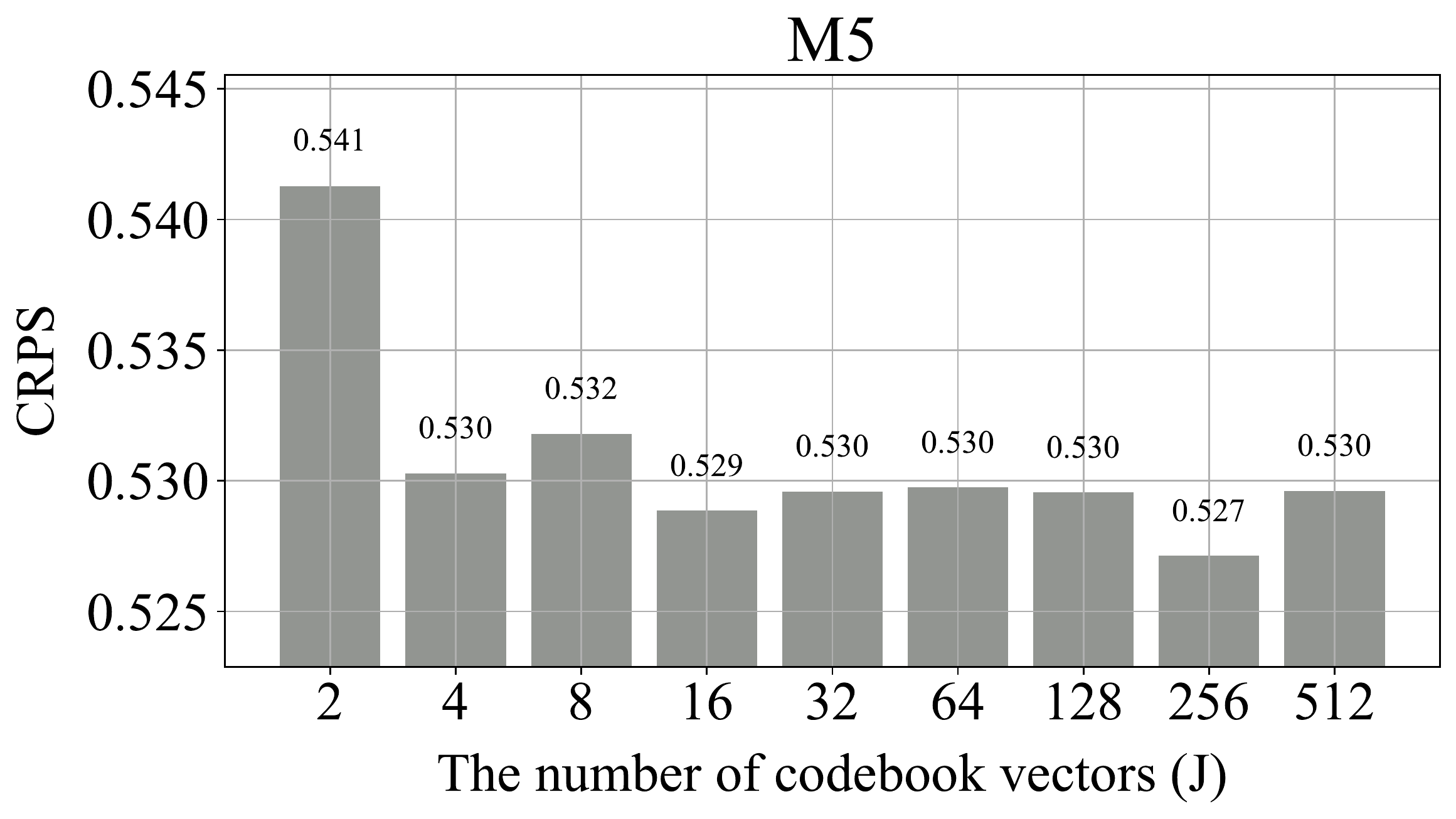}
        \caption{}
    \end{subfigure}
    \hfill
    \begin{subfigure}[b]{0.45 \textwidth}
        \centering
        \includegraphics[width=\textwidth]{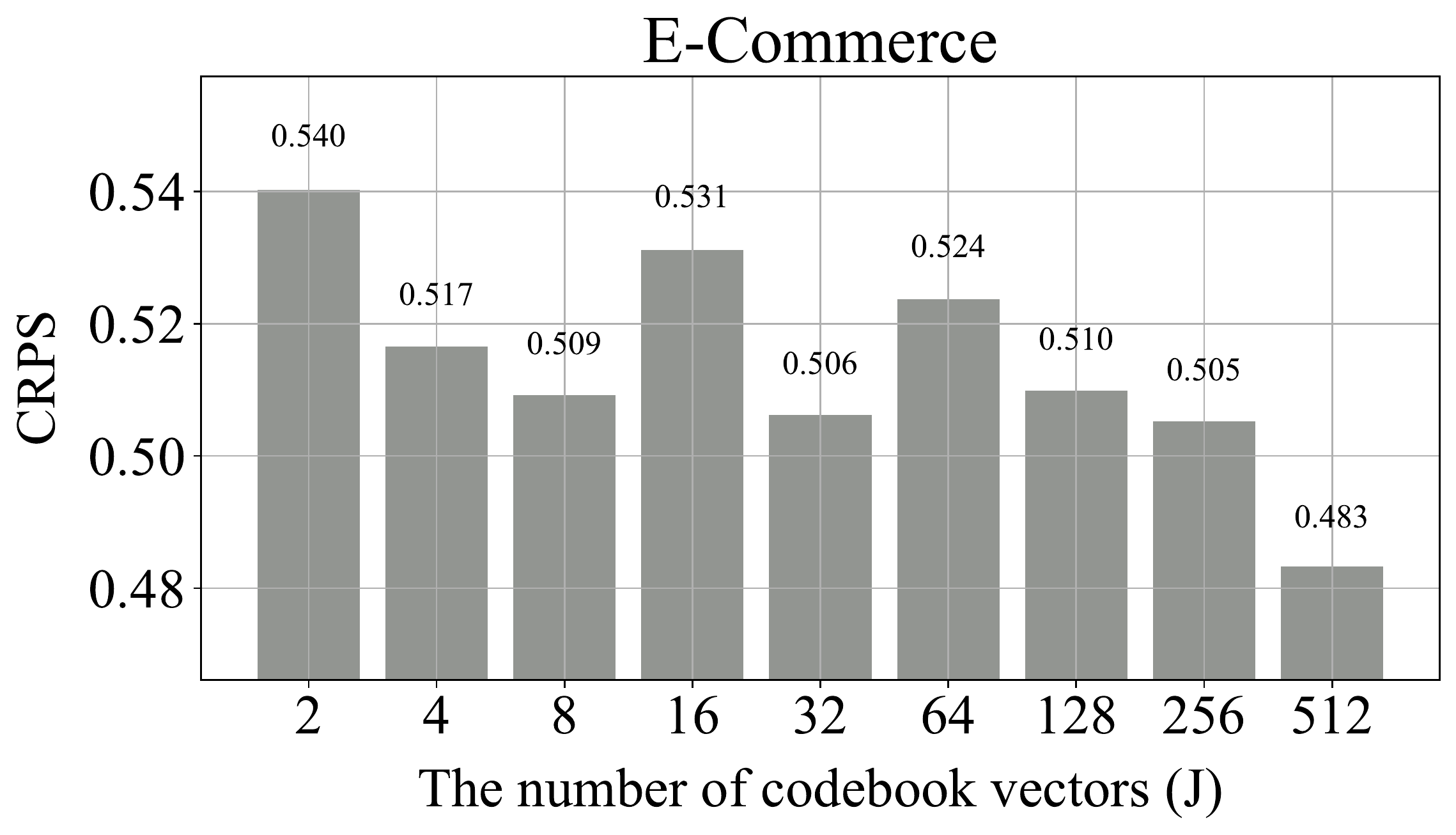}
        \caption{}
    \end{subfigure}
    \caption{Test set CRPS of \texttt{VQ-AR-iqn} on (a) \texttt{M5} and (b) \texttt{E-Commerce} datasets with different number of codebook vectors ($J$). Surprisingly, the decoding RNN can still give valid predictions given some sequence of just \emph{two} codebook vectors.}
    \label{fig:ablation_j}
\end{figure}

For our ablation study, we wish to test the performance of the model with respect to the number of codebook vectors $J$. In this experiment for the \texttt{M5} and \texttt{E-Commerce} datasets we train and test the performance of \texttt{VQ-AR} using $J= \{2, 4, \ldots, 512 \}$. We record the metrics in Figure~\ref{fig:ablation_j} and to our surprise, we see that we get decent forecasts using only \emph{two} codebook vectors for all the datasets considered (and not only for the ones in the figure). More codebook vectors  tend to give better performance. As $J \to \infty$ we approach the \texttt{DeepAR} method albeit with more RNN layers.

\section{Related Work} \label{sec:rel}

The use of Vector Quantization (VQ) in time series \emph{forecasting} has not been explored,  as far as we are aware. However, generative models like VQ-VAE have been used extensively  for   sequential modeling problems like for videos \cite{DBLP:conf/visapp/RakhimovVAZB21} and audio/speech \cite{dhariwal2020jukebox, 50491, 
Baevski2020vq-wav2vec} generation. Typically these models are trained in a two-stage process where a VQ-VAE model is first trained and then a sequential generative model is trained on top of the codebook vectors. The \texttt{VQ-Wav2Vec} \cite{Baevski2020vq-wav2vec} uses a causal convolution to first learn a representation of an audio signal some steps into the future for further downstream task. This model is not autoregressive, takes bounded audio inputs only and is not able to be used as a forecaster which would need to account for the inductive bias of scale and additional covariates.
%The recent text to image \texttt{DALL·E} \cite{pmlr-v139-ramesh21a} model is another such example which provides competitive zero-shot result in the multi-modal setting.

The \texttt{SOM-VAE} \cite{DBLP:conf/iclr/FortuinHLSR19} model is related in that it uses  VQ to  learn interpretable discrete representation of sequential data for \emph{clustering}. Also in the  paper \cite{Rabanser}, the authors discretize the \emph{inputs} to time series models via different binning transformations, as a way to mitigate the scaling issue discussed in section  \ref{sec:scaling}, and investigate  performance given this input transformation on different architectures.

Neural forecasting methods \cite{benidis2020neural} are related with the caveat that all of them, as far as we are aware, can be thought of as some component that learns a representation of the context window in a continuous fashion together with an emission component, be it point forecasting or probabilistic. CNN \cite{bai2018empirical} based methods can typically be replaced by RNNs or causal Transformers \cite{NIPS2017_7181} and most works try to introduce some time series or problem specific inductive bias into the architectural choices \cite{LIM20211748} or emission models \cite{NIPS2018_8004} or inputs \cite{Rabanser} rather than explicitly on the representations themselves.

\section{Summary and Discussion} \label{sec:summ}

We have presented \texttt{VQ-AR} a conceptually simple and novel yet powerful autoregressive time series forecasting method based on learning discrete historic representations of time series histories in order to forecast, trained in an end-to-end fashion. We have shown not only does it perform better or similar to the competitive methods considered, but is also  able to achieve good performance in comparison to ensembles of hundreds of tree-based models which typically dominate forecasting benchmarks, all via a single model. The architectural principle applied in this method can be adapted for multivariate forecasting and the underlying RNN based encoder-decoder can be replaced by a Transformer if so desired.

For future work, we would like to test the inductive bias introduced in this work for the multivariate setting and also incorporate it within the Transformer architecture especially for very long time series sequences. As mentioned in the appendix \ref{sec:grad}, the gradient estimator considered can also be replaced by its smooth-relaxation variant or by methods with less bias to help improve the VQ training. 

Finally, we have also highlighted a critique of \texttt{VQ-AR} especially in terms of its robustness to noisy inputs, and in the future, we would like to investigate  how to mitigate this potential issue. 
% Another avenue of investigation can be to learn hierarchical representations similar to the VQ-VAE2 \cite{NEURIPS2019_5f8e2fa1} method where  discrete global and local, or temporal and spatial hierarchies of time series can be considered.

\section*{Acknowledgments}
We wish to acknowledge and thank the authors and contributors of the many open source libraries that were used in this work, in particular: GluonTS ~\citep{gluonts_jmlr}, NumPy~\citep{harris2020array}, Pandas~\citep{reback2020pandas}, Matplotlib~\citep{Hunter:2007}, Jupyter~\citep{Kluyver2016jupyter} and  PyTorch~\citep{NIPS2019_9015}.

\bibliographystyle{plain}
\bibliography{references}

% %%
% %% If your work has an appendix, this is the place to put it.
\appendix
\section{Gradient Estimators} \label{sec:grad}

\begin{figure}
  \centering
  \includegraphics[width=0.8 \textwidth]{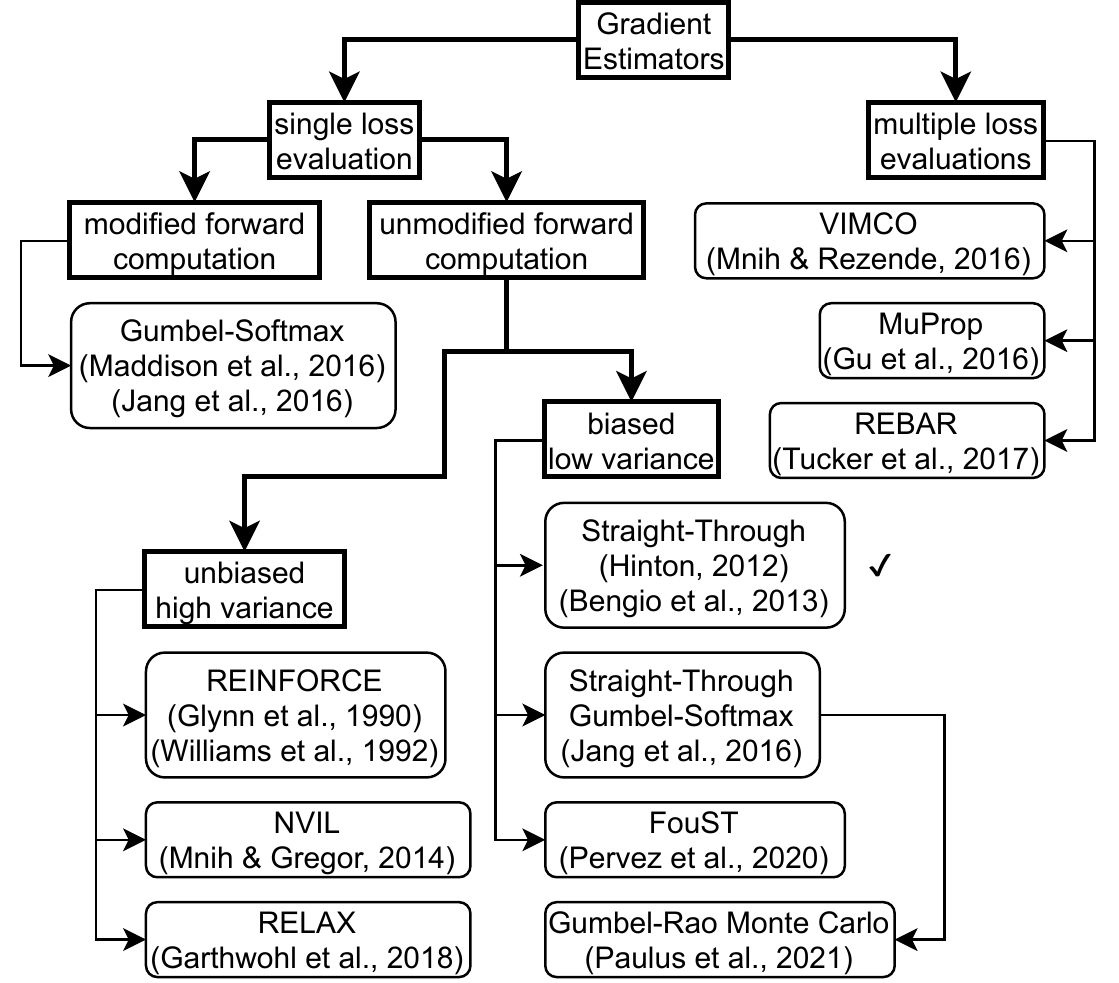}
  \caption{A taxonomy of  different gradient estimator methods, adapted from~\cite{paulus2021raoblackwellizing}, ordered chronologically with the method used in this paper marked with a tick.}
  \label{fig:gradient-estimators}
\end{figure}

Since the sampling and quantization in (\ref{eqn:posterior}) are not  differentiable the VQ-VAE paper applies a Straight-Through gradient estimator which propagates the gradients with respect to the codebook vectors through to the output of the encoding network. If the codebook vector and encoding representation are close then this is an adequate approximation but there is no bound on the bias of the gradients with this scheme and thus it falls under the biased-low variance taxonomy \cite{paulus2021raoblackwellizing} of gradient estimators as shown in Figure~\ref{fig:gradient-estimators}. 

\section{Experiment Details} \label{sec:exp-detail}

\begin{table}
\caption{Number of time series, domain, frequency, total training time steps and prediction length properties of the  training datasets used in the experiments.}
\label{dataset}
\begin{tabular}{l|ccccc}
\toprule
Dataset &   $D$ & Dom. & Freq. & Time  step  & Pred. len. %& CE
\\ 
\midrule
\texttt{Exchange}  & $8$ & $\mathbb{R}^{\geq 0}$ & day & $6,071$ & $30$ %& $16.86$
\\
\texttt{Solar}    & $137$ & $\mathbb{R}^{\geq 0}$ & hour & $7,009$ & $24$ %& $1535.38$
\\
 \texttt{Elec. } & $320$ &  $\mathbb{R}^{\geq 0}$ & hour & $15,782$ & $24$ %& $-21.97$
 \\
\texttt{Traffic}  & $862$  & $(0,1)$ & hour & $14,036$ & $24$ %& $68.62$
\\
\texttt{Taxi}    & $1,214$ & $\mathbb{N}^{\geq 0}$ & 30-min & $1,488$ & $24$ %& $-391.59$
\\
\texttt{E-Commerce} & $2,731$ & $\mathbb{N}^{\geq 0}$ & day& $887$ & $28$\\

\texttt{Wikipedia}   & $9,535$ & $\mathbb{N}^{\geq 0}$ & day & $762$ & $30$ %& $-1024.97$
\\
\texttt{M5} & $30,490$ & $\mathbb{N}^{\geq 0}$ & day & $1,941$ &   $28$ \\
\bottomrule
\end{tabular}
\end{table}

To help reduce confusion we also collect the different hyperparameters in a single location with their description and default values listed in Table~\ref{tab:notation}.

\begin{table}
    \caption{Glossary of the main hyperparameters, their description and default values unless specified.}
    \begin{tabular}{c|lc}
     \toprule
       Parameter  & Description  & Default \\
      \midrule
         $D$ & Number of time series in a dataset \\
         $P$ & Prediction horizon length \\
         $C$ & Context window length & $P \times 6$\\
         $F$ & Size of covariate vector  & \\
         $E$ &  \makecell[l]{Encoding RNN's hidden size and\\  size of codebook vectors} & 64\\
         $H$ & Decoding RNN's hidden size & 40\\
         $J$ & Number of codebook vectors & 128 \\
         $\beta$ & Commitment cost & $0.25$ \\
         $Q$ & Threshold to replace codebook vector & 2\\
         $S$ & Number of samples & 100\\
         
      \bottomrule
    \end{tabular}
    \label{tab:notation}
\end{table}

For our experiments we selected the default options for the models as implemented in the \texttt{GluonTS} \cite{gluonts_jmlr} library. In particular for \texttt{VQ-AR}, we modified the base implementation of \texttt{DeepAR} and thus kept the parameters of the models to be  similar where applicable. We trained each model for \texttt{max\_epochs=50} with the \texttt{batch\_size=256} and a default \texttt{learning\_rate=1e-3}. For the evaluation we used \texttt{quantiles=[0.1, 0.2, ..., 0.9]} to calculate the CRPS metric.

\end{document}